\documentclass[runningheads]{llncs}
\usepackage{graphicx}
\usepackage{subfigure}
\usepackage{amsmath}
\usepackage{amssymb}
\usepackage{wrapfig}
\usepackage{upgreek}
\begin{document}
\title{Truncated Gaussian-Mixture \\Variational AutoEncoder}
\titlerunning{Truncated Gaussian-mixture VAE}
%
\author{Qingyu Zhao\inst{1} \and
Nicolas Honnorat\inst{2} \and
Ehsan Adeli\inst{1} \and
Kilian M. Pohl\inst{1,2}}
\authorrunning{Zhao et al.}
%
\institute{Stanford University \and SRI International}

\maketitle              
\begin{abstract}
Variation Autoencoder (VAE) has become a powerful tool in modeling the non-linear generative process of data from a low-dimensional latent space. Recently, several studies have proposed to use VAE for unsupervised clustering by using mixture models to capture the multi-modal structure of latent representations. This strategy, however, is ineffective when there are outlier data samples whose latent representations are meaningless, yet contaminating the estimation of key major clusters in the latent space. This exact problem arises in the context of resting-state fMRI (rs-fMRI) analysis, where clustering major functional connectivity patterns is often hindered by heavy noise of rs-fMRI and many minor clusters (rare connectivity patterns) of no interest to analysis. In this paper we propose a novel generative process, in which we use a Gaussian-mixture to model a few major clusters in the data, and use a non-informative uniform distribution to capture the remaining data. We embed this truncated Gaussian-Mixture model in a Variational AutoEncoder framework to obtain a general joint clustering and outlier detection approach, called tGM-VAE. We demonstrated the applicability of tGM-VAE on the MNIST dataset and further validated it in the context of rs-fMRI connectivity analysis.


\end{abstract}
\section{Introduction}
Generative models in combination with neural networks, such as variational autoencoders (VAE), have gained tremendous popularity in learning complex distribution of training data by embedding them into a low-dimensional latent space. Traditional VAEs usually incorporates simple priors, e.g., a single Gaussian, for regularizing latent variables. Recently, in order to enhance the modelling capacity of VAE, mixture models have been used in the latent space to capture the multi-modal nature of the data and to perform unsupervised clustering. However, a challenging situation arising from some applications is that the key major clusters to be investigated are contaminated by many non-informative small clusters or by data corresponding to noise/outliers, whose latent representations are meaningless. For example, in the context of brain functional connectivity analysis, detecting major clusters of dynamic connectivity patterns is often hindered by the heavy noise of rs-fMRI (resting-state functional MRI) signals and many rare connectivity patterns of no interest to analysis. In the following, we will further setup this clustering problem for rs-fMRI analysis and then introduce a generic clustering method based on the VAE framework. The method uses a truncated Gaussian-Mixture model in the latent space to robustly capture the major clusters in the presence of outliers or minor clusters. 

Functional connectivity refers to the functionally integrated relationship between spatially separated brain regions \cite{Buckner13}. Recent work revealed that functional connectivity exhibits meaningful variations within the time series captured by resting-state fMRI \cite{Allen12,Zalesky14}. As a consequence, a considerable amount of work has been directed to quantify dynamic functional connectivity. A popular way of quantification \cite{Allen12,Damarajua14,Yu15} is to perform clustering on the time-varying connectivity patterns of a subject or a population. The resulting clusters then represent different functional connectivity states present in the data \cite{Damarajua14}.

Most existing works group the dynamic connectivity patterns into a fixed number clusters (k$\leq$5) to represent {\it major connectivity states} (commonly observed states) \cite{Allen12,Damarajua14,Yu15}. However, some studies have indicated that there exist many {\it minor states} containing rare connectivity patterns that persist shortly ($<1\%$ occupancy rate) \cite{Taghia17}. These minor states often provide little merit to analysis because they may correspond to random individual brain variation, inaccurate connectivity pattern computed during state transitions or rs-fMRI noise. Instead of merging minor states into the major ones \cite{Damarajua14,Yu15}, recent work suggests to disentangle minor from major states by modeling an infinite number of clusters \cite{Taghia17,Nielsen15}. However, connectivity patterns in minor states may correspond to pure noise, so grouping them into clusters is not meaningful. In this paper, we address these concerns by developing a statistical framework where the patterns associated with major states are drawn from an informative distribution while we use a non-informative distribution for minor states. 

Motivated by the truncated stick-breaking representation of Dirichlet processes \cite{Blei06,Nalisnick17}, our approach is guided by a Dirichlet prior that truncates the data into a few major clusters and a separate non-informative class. The major clusters are generated by a non-linear process from a low-dimensional latent space, where the latent representations follow a Gaussian-mixture distribution. The remaining data are generated from a uniform distribution in the original space. To determine the optimal parameters of our model, we derive the variational lower-bound of its log marginal probability and find the maximum of that lower-bound by optimizing a variational autoencoder. As a result, our method, tGM-VAE, simultaneously achieves clustering and outlier-detection.

In the context of functional connectivity analysis, it separates dynamic connectivity patterns into major and minor states. We define dynamic connectivity patterns by computing correlation matrices associated with sliding windows. The correlation matrices belonging to major states are associated with major clusters, and the rest of the correlation matrices, which correspond to minor states, are treated as outliers. 

In this work, we first apply tGM-VAE to the MNIST dataset as a proof-of-concept example. Then we demonstrate that tGM-VAE achieves higher accuracy in defining major clusters and outliers compared to traditional Gaussian-mixture-based approaches when clustering synthetic data with ground-truth. We then report that, for 15k correlation matrices derived from rs-fMRI scans of 593 adolescents in the the National Consortium on Alcohol and Neurodevelopment in Adolescence (NCANDA), tGM-VAE identifies meaningful connectivity states and a significant effect of age on their mean dwell time.


In the following, we first review existing VAE-based clustering approaches in Section 2. We introduce in Section 3 the generative model of tGM-VAE, the variational lower bound of the resulting log marginal likelihood, and reformulate tGM-VAE into a joint clustering and outlier-detection approach. A proof-of-concept example based on the MNIST dataset is discussed in Section 4. Finally, we present our experiments on rs-fMRI data analysis in Section 5.

\section{Related Work} 
Traditional clustering approaches are mostly based on Gaussian-mixture models \cite{Damarajua14,Yu15}. These methods usually require fitting probability distributions in a high dimensional space, which is a challenging task. Moreover, it has been found that the underlying distributions of both fMRI measurements \cite{Taghia17} and the derived correlation matrices \cite{Zhao18} lie on a non-linear latent space. Therefore, modeling Gaussian-mixtures in the original space is suboptimal. 

Generative models used in connection with neural-networks, such as VAEs, have recently attracted much attention for their capability of modeling latent representations of the data \cite{Kingma13}. In VAE, the encoder approximates the intractable posterior distribution of the latent representation and the decoder aims to reconstruct the observation based on its latent representation. While traditional VAE assumes that latent variables follow a single Gaussian prior, recent works adopt mixture models in the latent space for semi-supervised learning \cite{Kingma14} and clustering \cite{Dilokthanakul17}. Dilokthanaku et al. \cite{Dilokthanakul17} construct a two-level latent space that allows for a multi-modal prior of latent variables, but this model exhibits over-regularization effects that require specific optimization procedures. Jiang et al. \cite{Jiang17} explicitly define a generative process based on a mixture of Gaussians in the latent space, which achieves better clustering performance. Our tGM-VAE model is built upon a generative model similar to \cite{Jiang17} to capture major states but also includes a non-informative distribution for modeling minor states.

Besides the above approaches for modeling fixed number of clusters, Bayesian non-parametric models have been adopted to model an infinite number of clusters. The semi-supervised approach proposed in \cite{Abbasnejad17} uses multiple VAEs as a proxy of Gaussian-mixture models and automatically determines the number of VAEs by maximizing the reconstruction capability for the entire dataset. The stick-breaking construction \cite{Nalisnick17} has also been adopted in VAE for semi-supervised classification, where the latent representation is a set of truncated categorical weights. While this approach is not intrinsically built for clustering, the truncation strategy motivates us to use the last category (remainder of the truncation) to capture all dynamic connectivity patterns that do not belong to major clusters. Contrary to the above two approaches, tGM-VAE only models the encoding/decoding process for major clusters and omits the latent representation for the remainder. This strategy is useful when the remainder corresponds to (a) minor clusters of no interest to analysis so modeling their latent presentations is redundant; (b) outliers whose latent representations are meaningless or do not form clusters.

\section{Methods}
\subsection{The Generative Model}
Let $\textbf{X}=\{\boldsymbol{x}_1,...,\boldsymbol{x}_N\}$ be a training dataset with $N$ observations. For example, each $\boldsymbol{x}_i$ represents a dynamic connectivity pattern, i.e., the upper triangular part of an ROI-to-ROI correlation matrix derived from the rs-fMRI time series at a given sliding window \cite{Allen12}. We assume that each $\boldsymbol{x}_i$ belongs to a state, which, in our proposed generative process, is encoded by the categorical variable $c_i$. The first $K-1$ categories represent the major states and $c_i=K$ corresponds to the remainder (minor states). $c_i$ is drawn from a categorical distribution $p_{\boldsymbol{\pi}}(c_i)\sim\mbox{Cat}(\boldsymbol{\pi})$, where  $\boldsymbol{\pi}=[\pi^1,...,\pi^{K}]$
belongs to the $(K-1)$-dimensional simplex and is 
generated from a Dirichlet prior with two parameters $p(\boldsymbol{\pi})\sim\mbox{Dir}(\alpha,...,\alpha,\beta)$. By construction, a single parameter $\alpha$ controls the portion of the $K-1$ major clusters indifferently, and $\beta$ separately controls for the portion of the remainder via a stick-breaking procedure $\mbox{Beta}((K-1)\cdot \alpha,\beta)$ \cite{Blei06}. 

\begin{figure}[!t]
	\centering
    \subfigure[]{\includegraphics[width=0.3\linewidth]{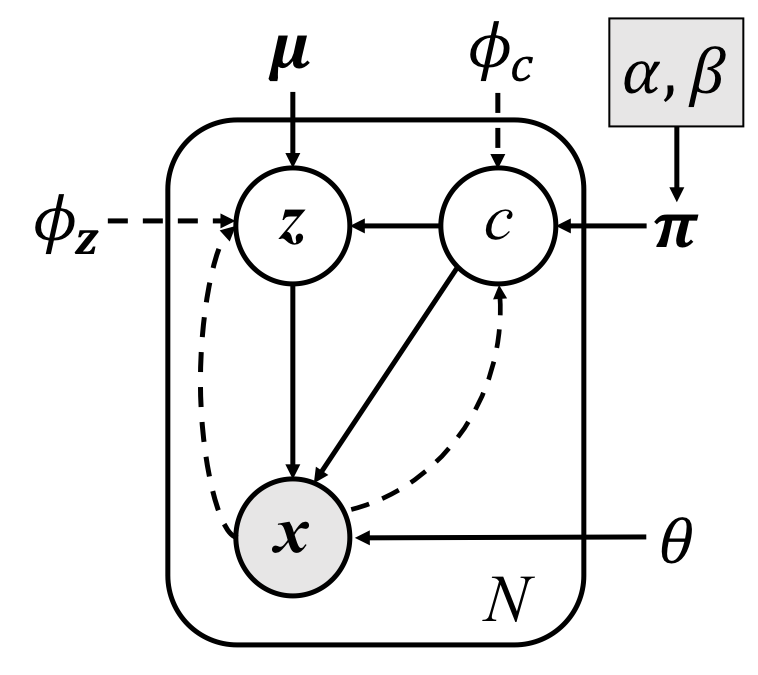}}
    \subfigure[]{\includegraphics[width=0.3\linewidth]{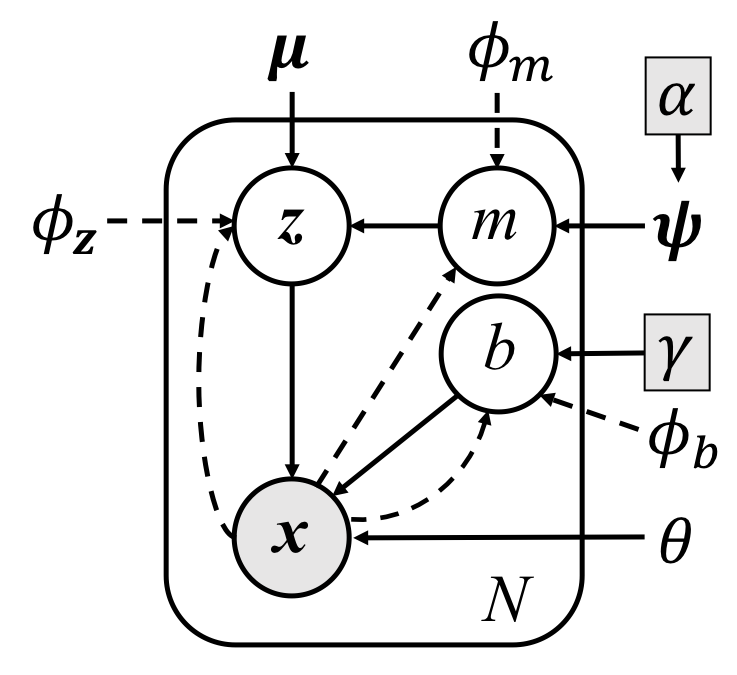}}
    \subfigure[]{\includegraphics[width=0.36\linewidth]{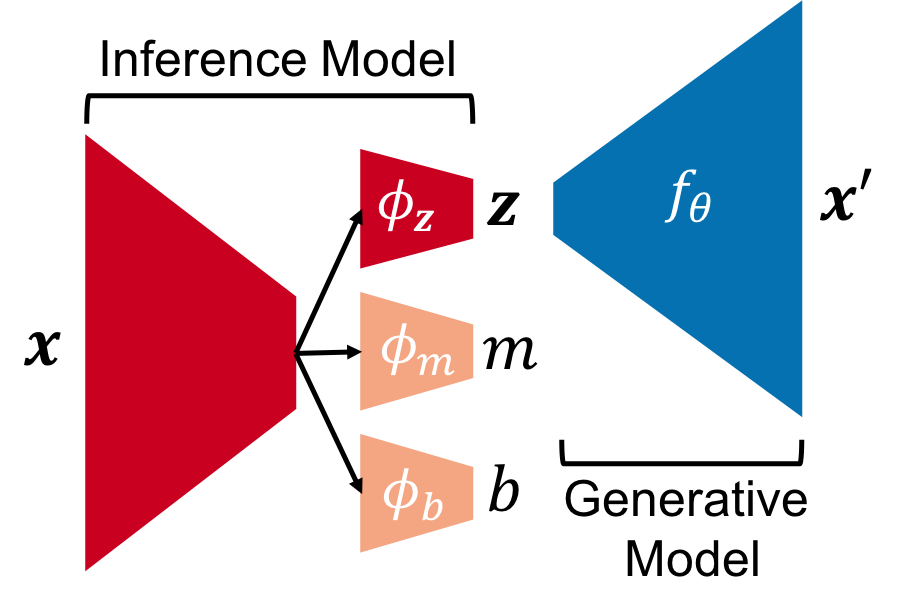}}
	\caption{(a) Bayesian model associated with tGM-VAE. Solid lines denote the generative model and dashed lines the inference model. Gray nodes denote observed variables and given parameters. (b) Reformulated model. (c) Neural network implemented from (b).}
	\label{fig:formulation}
\end{figure}

For simplicity, let $c^k_i$ denote $c_i=k$. We assume that when $c^k_i$ with $k<K$, $\boldsymbol{x}_i$ is generated from a latent representation $\boldsymbol{z}_i$ through a non-linear process modeled by a neural-network $f$ with parameter $\theta$: $p_{\theta}(\boldsymbol{x}_i|\boldsymbol{z}_i)\sim \mathcal{N}(f_\theta(\boldsymbol{z}_i),\sigma^2_x)$, where $\sigma^2_x$ is the fixed standard deviation of noise. We further assume $\boldsymbol{z}_i$ is drawn from a Gaussian distribution with mean $\boldsymbol{\mu}_k$ and an identity covariance: $p_{\boldsymbol{\mu}}(\boldsymbol{z}_i|c^k_i) \sim \mathcal{N}(\boldsymbol{\mu}_k,\textbf{I})$ with $\boldsymbol{\mu}=\{\boldsymbol{\mu}_k|k<K\}$. In other words, the marginal distribution of $\boldsymbol{z}_i$ follows a Gaussian mixture in the latent space. On the other hand, when $c_i=K$, we assume $\boldsymbol{x}_i$ is simply drawn from a uniform distribution in a unit domain $\xi$ embedded in the original space containing all observations after normalization: $p(\boldsymbol{x}_i|c^K_i)\sim\mathcal{U}(\xi)$. Based on the above generative model parameters $\Theta=\{\boldsymbol{\pi},\boldsymbol{\mu},\theta\}$, we have $p_{\Theta}(\boldsymbol{x}_i,\boldsymbol{z}_i,c_i^k)=p_{\theta}(\boldsymbol{x}_i|\boldsymbol{z}_i)p_{\boldsymbol{\mu}}(\boldsymbol{z}_i|c^k_i)p_{\boldsymbol{\pi}}(c^k_i)$ for $k<K$, and $p_{\Theta}(\boldsymbol{x}_i,c_i^K)=p_{\boldsymbol{\pi}}(c^K_i)$. The Bayesian graphical diagram of this model is given in Fig. \ref{fig:formulation}a.

\subsection{Variational Lower Bound}
Given the training dataset $\textbf{X}$ and the two parameters $\{\alpha,\beta\}$ of the Dirichlet prior, the generative model parameters $\Theta$ are determined by maximizing the marginal probability $ p(\textbf{X},\boldsymbol{\pi}|\boldsymbol{\mu},\Theta,\alpha,\beta)$. Assuming i.i.d for each $\boldsymbol{x}_i$, the log marginal probability can be written as:
\begin{equation}
    \log p(\textbf{X},\boldsymbol{\pi}|\boldsymbol{\mu},\theta,\alpha,\beta)=\sum_{i=1}^N \log p_{\Theta}(\boldsymbol{x}_i) + \log p(\boldsymbol{\pi}|\alpha,\beta)
\end{equation}
In the above equation, the log likelihood $\log p_{\Theta}(\boldsymbol{x}_i)$ can not be directly optimized, so variational inference is used to maximize its lower-bound. Typically, lower-bounds for graphical models are derived by approximating an intractable posterior $p(\boldsymbol{z}_i,c_i|\boldsymbol{x}_i)$ on the latent variables with a tractable function $q(\boldsymbol{z}_i,c_i|\boldsymbol{x}_i)$. Here we make the common mean-field assumption: $q(\boldsymbol{z}_i,c_i|\boldsymbol{x}_i)=q(\boldsymbol{z}_i|\boldsymbol{x}_i)q(c_i|\boldsymbol{x}_i)$. When omitting the subscripts $i$ to simplify notations, it reads:
\begin{align}
    \mbox{log }p_{\Theta}(\boldsymbol{x}) &= \log\left(p_{\Theta}(\boldsymbol{x},c^{K})+\sum_{k=1}^{K-1}{ \int_{\boldsymbol{z}}{ p_{\Theta}(\boldsymbol{x},\boldsymbol{z},c^k)}}\right) \\
    &= \log\left(p_{\Theta}(\boldsymbol{x},c^K)\frac{q(c^K|\boldsymbol{x})}{q(c^K|\boldsymbol{x})}+\sum_{k=1}^{K-1}{\int_{\boldsymbol{z}}{ p_{\Theta}(\boldsymbol{x},\boldsymbol{z},c^k)\frac{q(\boldsymbol{z},c^k|\boldsymbol{x})}{q(\boldsymbol{z},c^k|\boldsymbol{x})}}} \right)\\
    &=\log\left(q(c^K|\boldsymbol{x})\frac{p_{\boldsymbol{\pi}}(c^K)}{q(c^K|\boldsymbol{x})}+\sum_{k=1}^{K-1}{q(c^k|\boldsymbol{x})\mathbb{E}_{q(\boldsymbol{z}|\boldsymbol{x})}\left[\frac{p_{\Theta}(\boldsymbol{x},\boldsymbol{z},c^k)}{q(\boldsymbol{z},c^k|\boldsymbol{x})}\right]}\right)\\
    &\geq q(c^K|\boldsymbol{x})\log\frac{\pi^K}{q(c^K|\boldsymbol{x})}+\sum_{k=1}^{K-1}{ q(c^k|\boldsymbol{x})\mathbb{E}_{q(\boldsymbol{z}|\boldsymbol{x})}\left[\log\frac{p_{\Theta}(\boldsymbol{x},\boldsymbol{z},c^k)}{q(\boldsymbol{z},c^k|\boldsymbol{x})}\right]} \\
    &= \sum_{k=1}^{K}{q(c^k|\boldsymbol{x})\log\frac{\pi^k}{q(c^k|\boldsymbol{x})}}+\sum_{k=1}^{K-1}{q(c^k|\boldsymbol{x})\mathcal{L}^k(\boldsymbol{x})},\mbox{ where}\\
    \mathcal{L}^k(\boldsymbol{x}) &:=\mathbb{E}_{q(\boldsymbol{z}|\boldsymbol{x})}\left[p_{\theta}(\boldsymbol{x}|\boldsymbol{z})\right]-D_{\mbox{KL}}\left(q\left(\boldsymbol{z}|\boldsymbol{x}\right)~||~p_{\boldsymbol{\mu}}\left(\boldsymbol{z}|c^k\right)\right)
    \label{eq:vlb}
\end{align}
where $D_{\mbox{KL}}$ denotes the KL divergence between two probability distributions. 
~\\~\\
\textbf{Interpretation of the Lower Bound.} $\mathcal{L}^k$ corresponds to the formulation of the traditional single-Gaussian VAE \cite{Kingma13} with respect to the $k^{th}$ cluster. Specifically, $\mathbb{E}_{q(\boldsymbol{z}|\boldsymbol{x})}[p_{\theta}(\boldsymbol{x}|\boldsymbol{z})]$ encourages the decoded reconstruction of the latent variable to resemble the observation. The $D_{\mbox{KL}}$ term is commonly interpreted as a regularizer encouraging the approximate posterior $q(\boldsymbol{z}|\boldsymbol{x})$ to resemble the cluster-specific Gaussian prior $p_{\boldsymbol{\mu}}(\boldsymbol{z}|c^k)$. 

The right term of the lower-bound (Eq. 6) sums the losses of single-Gaussian VAEs over the $K-1$ major clusters and weighs them by cluster-assignment probability $q(c|\boldsymbol{x})$. Maximizing this term improves the encoding/decoding capability for patterns in major states while keeping their latent variables to form clusters. The left term of the lower-bound corresponds to the KL-divergence between $q(c|\boldsymbol{x})$ and $\mbox{Cat}(\boldsymbol{\pi})$ and encourages the posterior categorical distribution to approximate the categorical prior. It is important to note that latent representations are only modeled for the $K-1$ clusters but not for the remainder. The portion of the remainder is controlled by the left term of the lower-bound.

\subsection{Reformulation} 
In this section, we reformulate our model to demonstrate, by re-organizing the lower-bound of Eq. 6, that tGM-VAE can be interpreted as a joint outlier-detection and clustering framework. Given the generative process described in Section 3.1, the categorical variable $c$ can be constructed by first differentiating the major clusters from the remainder. Let $b$ denote a Bernoulli variable generated by $p(b) \sim \mbox{Ber}(\gamma)$, where $\gamma \in [0,1]$ defines the portion of the remainder. When $b^0$ ($b=0$ for major clusters), a cluster assignment variable $m$ is drawn from a categorical distribution $p_{\boldsymbol{\psi}}(m) \sim \mbox{Cat}(\boldsymbol{\psi})$, where $\boldsymbol{\psi}=[\psi^1,...,\psi^{K-1}]$ follows $ \mbox{Dir}(\alpha,...,\alpha)$. This construction also involves two parameters, $\{\alpha,\gamma\}$. The graphical diagram of this model is given in Fig. \ref{fig:formulation}b.

For posterior inference, different $q$ functions are constructed for the reformulated generative process. Let $q(b|\boldsymbol{x})$ denote the approximate posterior of assigning $\boldsymbol{x}$ to either major clusters or the remainder and let $q(m|\boldsymbol{x},b^0)$ denote the major cluster assignment given $b^0$. Then $q(c|\boldsymbol{x})$ and $\boldsymbol{\pi}$ in Section 3.2 become 
\begin{align}
    q(c^k|\boldsymbol{x})=q(m^k|\boldsymbol{x},b^0)q(b^0|\boldsymbol{x}) &\mbox{ for } k<K, \mbox{ and } q(c^K|\boldsymbol{x})=q(b^1|\boldsymbol{x})\\
    \pi^k = \psi^k(1-\gamma) &\mbox{ for } k<K, \mbox{ and } \pi^K = \gamma
    \label{eq:outlier}
\end{align} 
Replacing the terms in Eq. 6 with Eq. 8,9 leads to the following lower bound
\begin{equation}
    \mbox{log }p_{\Theta}(\boldsymbol{x}) \geq q\left(b^0|\boldsymbol{x}\right) \mathcal{G}(\boldsymbol{x},m)-D_{\mbox{KL}}\left(q\left(b|\boldsymbol{x}\right)~||~\mbox{Ber}\left(\gamma\right)\right),
    \label{eq:vlb_outlier}
\end{equation}
where $\mathcal{G}(\boldsymbol{x},m)=\sum_{k=1}^{K-1} q(m^k|\boldsymbol{x},b^0)\mathcal{L}^k(\boldsymbol{x})-D_{\mbox{KL}}(q(m|\boldsymbol{x},b^0)||\mbox{Cat}(\boldsymbol{\psi}))$ is exactly the formulation of Gaussian-mixture VAE with $K-1$ clusters \cite{Jiang17}. From Eq. \eqref{eq:vlb_outlier} we can see that $q(b^0|\boldsymbol{x})$ essentially gives the probability of $\boldsymbol{x}$ being an inlier. Data with high inlier-probability are then clustered by $\mathcal{G}(\boldsymbol{x},m)$, while the right term in Eq. \eqref{eq:vlb_outlier} regularizes the portion of outliers with parameter $\gamma$. In practice, we use an additional weight $\lambda$ to balance the two types of losses in Eq. \eqref{eq:vlb_outlier}, a common practice in VAE frameworks \cite{Dilokthanakul17,Higgins17}.

\subsection{Network Design}
The design of our VAE network is based on the above inference procedure. More specifically, all the approximate posteriors are modeled by neural networks. Similar to the traditional VAE \cite{Kingma13}, $q(z|x)$ is an encoder network (Fig. \ref{fig:formulation}c red blocks) with parameters $\phi_z$, which encodes the posterior as a multivariate Gaussian with an identity covariance $q(\boldsymbol{z}|\boldsymbol{x})=\mathcal{N}(\boldsymbol{z};\Tilde{\mu},\textbf{I})$. While allowing for a diagonal or full covariance are both reasonable practices, we simply rely on the non-linear neural network to capture the covariance structure, and we only use the mean to capture the clustering effects in the latent space. The encoder has 3 densely connected hidden layers  with \textit{tanh} activation. The dimensions of the 3 layers will be introduced in the following sections on experiments. The decoder network $f_\theta(\boldsymbol{z})$ has an inverse structure as the encoder and uses MSE reconstruction loss. For the optimization of these two networks, the SGVB estimator and reparameterization trick are adopted \cite{Kingma13}.  

Contrary to previous work \cite{Jiang17,Ebbers17}, we also use neural networks to model the categorical posteriors $q(b|\boldsymbol{x})$ and $q(m|\boldsymbol{x})$ (Fig. \ref{fig:formulation}c orange blocks). Their first two layers were shared from the encoder of $q(\boldsymbol{z}|\boldsymbol{x})$ and the last layer is densely connected with \textit{soft-max} activation. This construction rigorously reflects the structure of the generative model described in Section 3.3 (Fig. \ref{fig:formulation}b) and allows for two separate mechanisms for detecting outliers with $q(b|\boldsymbol{x})$ and assigning clusters with $q(m|\boldsymbol{x})$. By comparison, a single neural network for $q(c|\boldsymbol{x})$ would be obtained from the model described in Section 3.1 (Fig. \ref{fig:formulation}a), but this network would treat the clusters and outliers indifferently. 

\section{Proof-of-Concept on MNIST}
\begin{figure}[!b]
	\centering
    \includegraphics[width=1\linewidth]{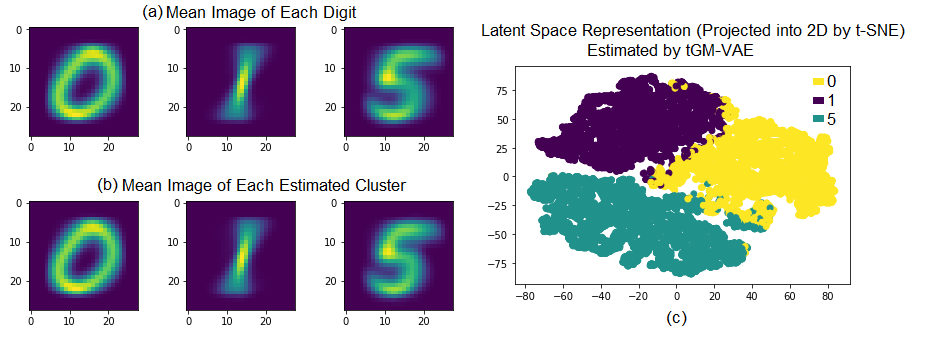}
	\caption{(a) Mean images of 3 digits that were randomly selected out of 10. (b) Mean images associated with the 3 major clusters estimated by tGM-VAE. (c) 2D visualization of latent representations of the 3 major clusters.}
	\label{fig:MNIST}
\end{figure}

3 digits out of 10 were randomly selected, and the corresponding images associated with the 3 digits in the MNIST dataset were treated as data of major clusters (Fig. \ref{fig:MNIST}a). We then randomly sampled images of the remaining 7 digits from MNIST, such that the 3 major clusters composed of 90\% of the our final training data. The dimensions of the 3 layers in the encoder are $(384,64,4)$. The decoder network $f_\theta(\boldsymbol{z})$ has an inverse structure as the encoder and uses binary cross-entropy loss. For this experiment, we used the following parameter settings: $\gamma=0.1$, $\beta=1.1$, and $\lambda = 200$. These settings corresponded to an accurate estimate of the portion of the remainder ($\gamma$), a rather non-informative Dirichlet prior ($\beta$) and a strong regularization on the portion of the remainder ($\lambda$).

The ground-truth labels were defined with respect to 4 classes: 3 major clusters and the remainder. The 4-class accuracy (percentage of all data samples that were labelled correctly \cite{Jiang17}) achieved by tGM-VAE was 88\%. The 3-class accuracy (accuracy w.r.t. to the 3 selected digits) was 92\%. 54\% of the remaining 7 digits were labelled as outliers. The mean image of each estimated cluster and the estimated latent space are displayed in Fig. \ref{fig:MNIST}. 

\section{Experiments on rs-fMRI Data}
tGM-VAE was first validated and compared to traditional clustering approaches based on synthetic experiments, where rs-fMRI series and time-varying correlation matrices were simulated according to a ground-truth state sequence. We measured, in particular, the accuracy of tGM-VAE in connectivity states estimation. Then, tGM-VAE was used to cluster 15k correlation matrices obtained from the rs-fMRI scans of 593 adolescents in the NCANDA study \cite{MO17}. The relation between the age of a subject and the mean dwell time of the connectivity states was finally examined.

\subsection{Synthetic Experiments}
\textbf{Data Simulation.} We followed the simulation procedure presented in \cite{Allen12,Taghia17}  by first generating a state sequence of 50000 time points associated with 10 connectivity states, among which 5 states were major states. The transition probability from the $i^{th}$ state to the $j^{th}$ state was set to $0.9\delta_{ij}+0.1b_i^j$, where $\delta$ is the Kronecker Delta function, and $\boldsymbol{b_i}=[b_i^1,...,b_i^{10}]$ was randomly generated from $\mbox{Dir}(10,...,10,1,...,1)$. This process led to self-transition probabilities varying between 0.9 and 0.95, and cross-state transition probabilities between 1e-4 and 0.05. The mean dwell time of a state (average time that a state continuously persists before switching to another state) varied between 8 and 15 time points. The occupancy rate of a major state (percentage of a state occupying the sequence) varied between 8\% to 30\%, and the total occupancy of the 5 minor states varied between 5\% to 10\%. These metrics are similar for real rs-fMRI data reported in \cite{Taghia17}.

\begin{figure}[!b]
	\centering
    \includegraphics[width=0.95\linewidth]{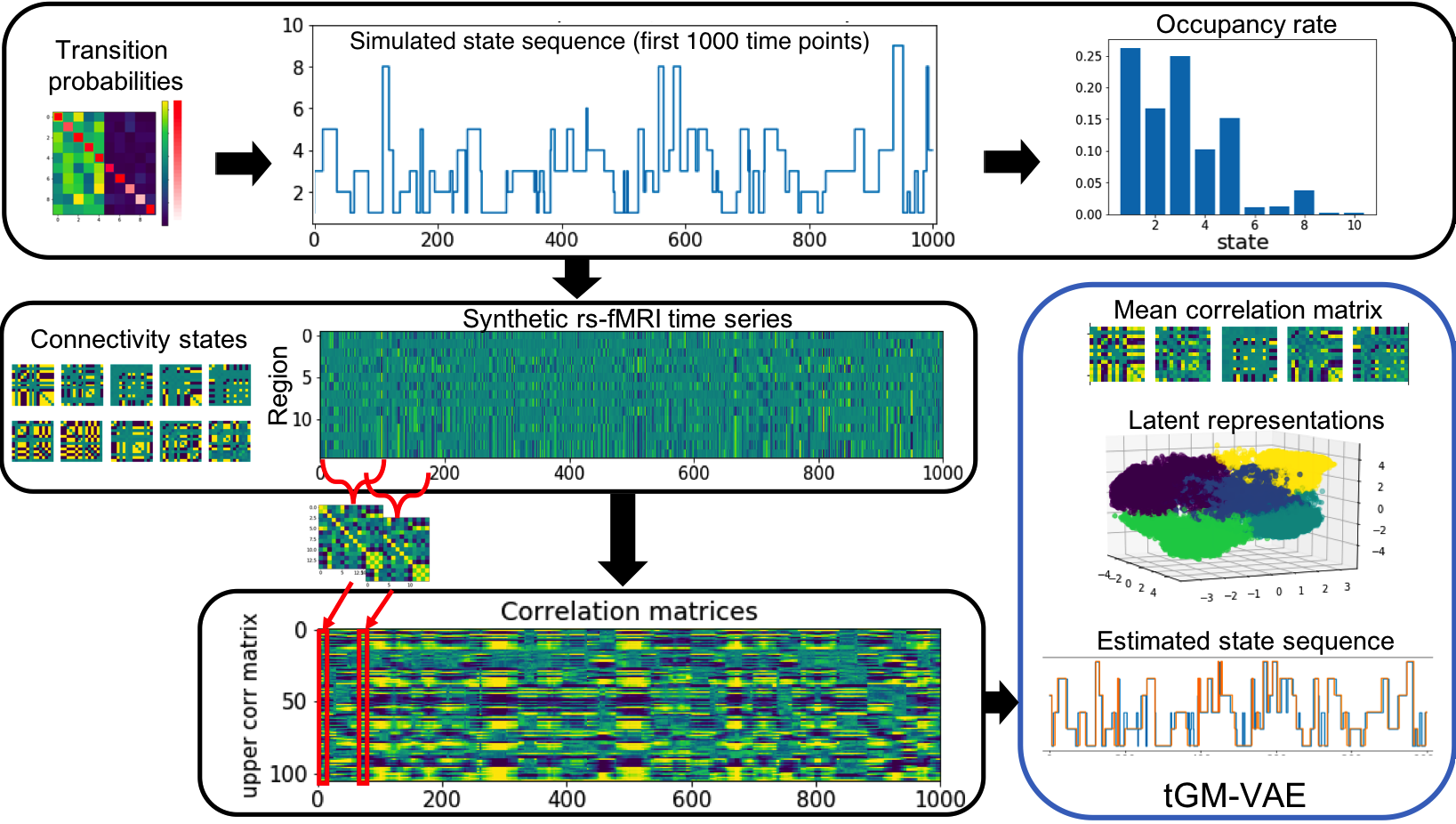}
	\caption{Pipeline for simulating synthetic data.}
	\label{fig:synthetic_setup}
\end{figure}

Next, a connectivity pattern was simulated for each state. In the first experiment, we assumed that there were 15 regions of interest (ROI) in the brain, so each state was associated with a $15\times15$ matrix, known as the community matrix \cite{Taghia17}. For the $i^{th}$ state, a 1D loading vector $\boldsymbol{u}_i\in\mathbb{R}^{15}$ consisted of $\{1,-1,0\}$ (representing positive/negative or no activation of each ROI) was randomly generated. Then, the $i^{th}$ community matrix was computed by $\boldsymbol{u}_i\boldsymbol{u}_i^\textbf{T}$ \cite{Allen12}.

Afterwards, synthetic rs-fMRI signals at each time point were randomly sampled from a Gaussian distribution with the covariance being the state-specific community matrix at that time point. Gaussian noise of standard deviation 0.1 was further added to the synthetic rs-fMRI series. Finally, dynamic correlation matrices were generated using a sliding window of length 11. These different steps are summarized in Fig. \ref{fig:synthetic_setup}.

\textbf{Clustering Accuracy.} tGM-VAE clustered the dynamic correlation matrices into 5 major states with the following parameter settings: $\gamma=0.075$, $\beta=1.1$, and $\lambda = 200$. These settings corresponded to an accurate estimate of the portion of the remainder ($\gamma$), a rather non-informative Dirichlet prior ($\beta$) and a strong regularization on the portion of the remainder ($\lambda$). The dimensions of the 3 layers in the encoder were $(D,16,3)$, where $D$ is the leading ``power of two'' that is smaller than the input dimension (e.g., D=64 for a 15$\times$15 correlation matrix with 105 upper triangular elements).

\begin{figure}[!b]
	\centering
    \subfigure[]{\includegraphics[width=0.45\linewidth]{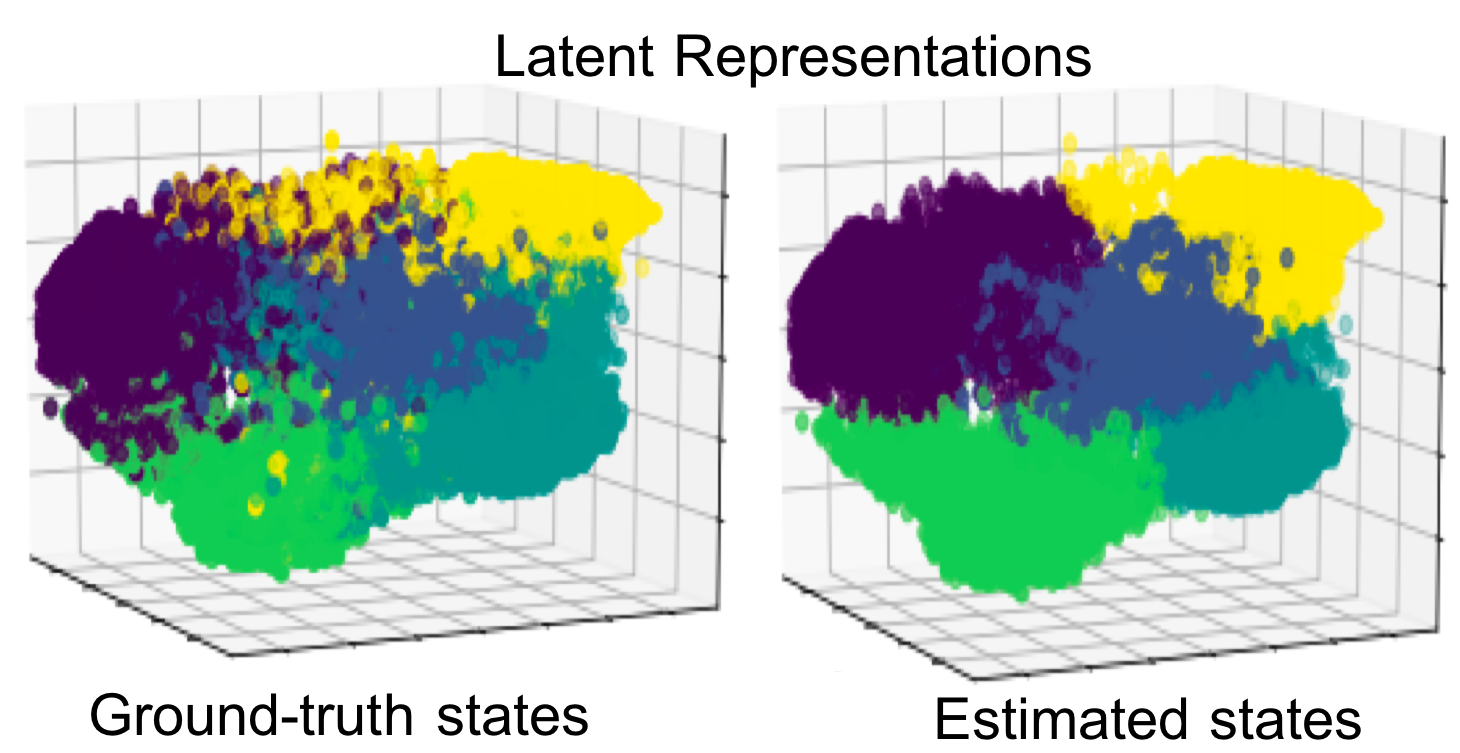}}
    \subfigure[]{\includegraphics[width=0.53\linewidth]{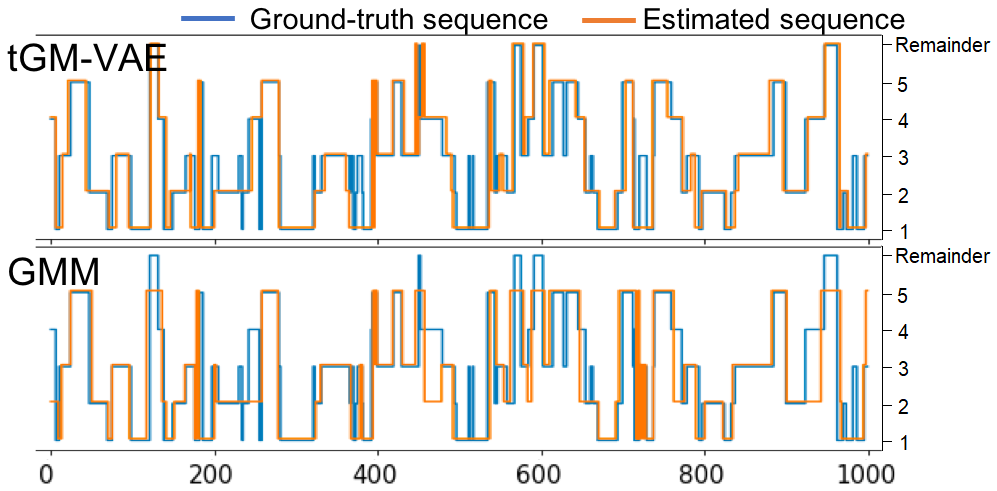}}
	\caption{(a) Latent representations of correlation matrices computed by tGM-VAE color-coded by ground-truth states (left) and estimated states (right). (b) State sequences estimated by tGM-VAE and GMM overlaid with the ground-truth sequence. }
	\label{fig:tgm-vae}
\end{figure}

Fig. \ref{fig:tgm-vae} presents the 3D latent space associated with tGM-VAE. Only the 5 major states are displayed as the latent representations of the remainder were not modeled. We can observe that the latent representations were reasonably clustered by states, thanks to the Gaussian-mixture modeling in the latent space \cite{Nalisnick17,Dilokthanakul17,Jiang17}. 

\begin{figure}[!t]
	\centering
    \includegraphics[width=1\linewidth]{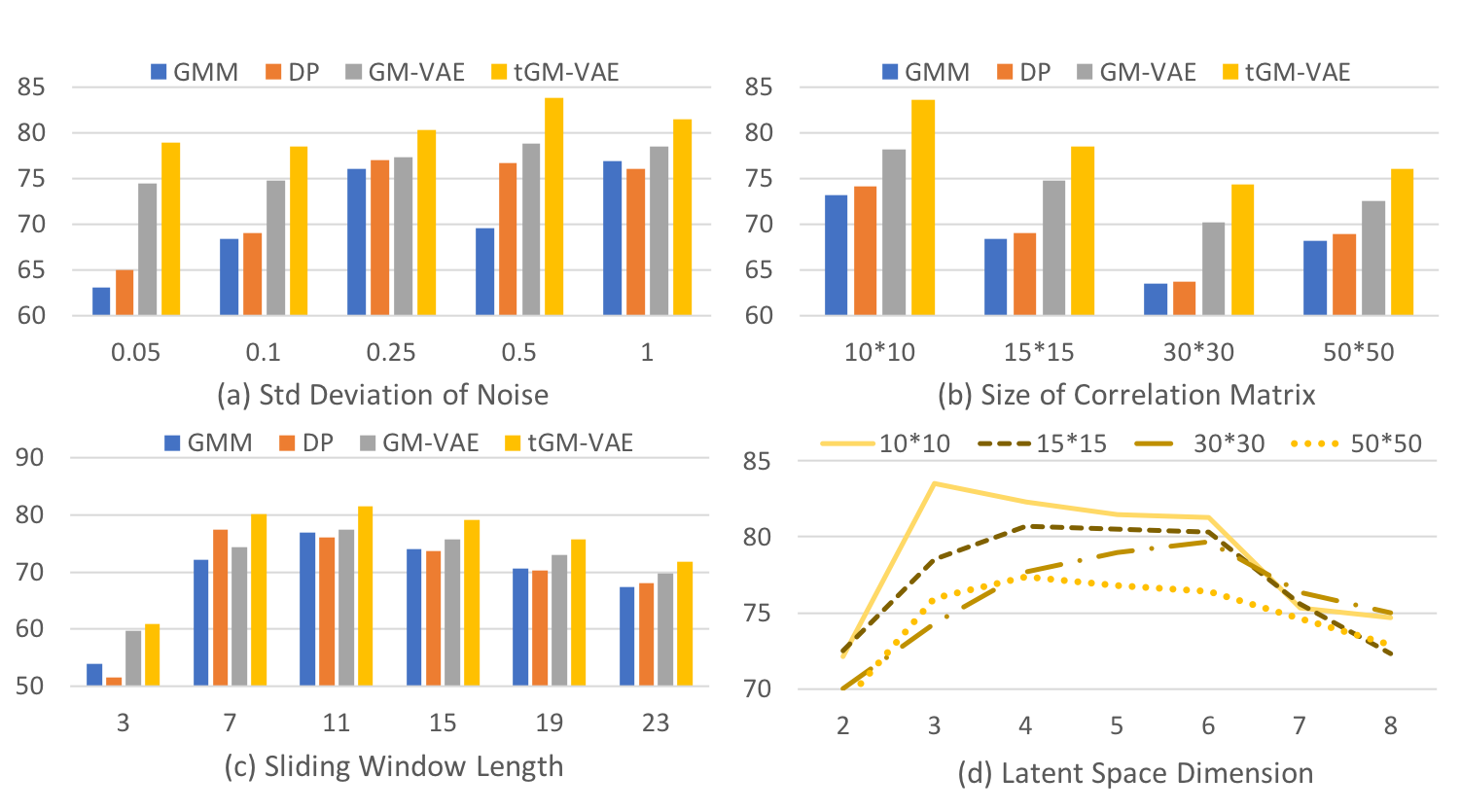}
	\caption{Clustering accuracy scores measured on synthetic data by varying (a) noise level; (b) size of correlation matrix; (c) sliding window length. (d) tGM-VAE accuracy as a function of latent space dimension.}
	\label{fig:various_settings}
\end{figure}

To associate the 5 estimated clusters with the 5 ground-truth major states, the correlation matrices in an estimated cluster were first averaged and linked to the closest community matrix with respect to the Frobenius norm. As there was no interest in differentiating minor connectivity states, the clustering accuracy was measured with respect to the 6 classes (5 clusters + remainder).  tGM-VAE was compared with three other clustering approaches as indicated by Fig. \ref{fig:various_settings}. Both Gaussian-Mixture Model (GMM) and Gaussian-Mixture VAE (GM-VAE) clustered the entire dataset into 5 clusters (merging minor states into major ones); The non-parametric Dirichlet Process (DP) Gaussian-mixture approach modeled an infinite number of clusters, so the 5 largest clusters estimated by DP were considered major states and the rest was considered the remainder. The clustering accuracy of these approaches was 68.4\% (GMM), 69.0\% (DP), 74.8\% (GM-VAE) and 78.5 \% (tGM-VAE).  Fig. \ref{fig:tgm-vae}b shows the estimated state sequence produced by tGM-VAE (most accurate) and GMM (least accurate). We observe that the two VAE-based methods produced significantly improved clustering accuracy than the two traditional Gaussian-Mixture methods (GMM and DP). This improvement indicates that the modeling of latent representations and the associated non-linear generative processes as provided by the VAE framework were helpful in analyzing correlation matrices. Moreover, the truncation of tGM-VAE could accurately capture the minor states and provided 3.7\% improvement over GM-VAE, whereas explicitly clustering minor states was a less effective strategy (DP only 0.6\% improvement over GMM).

Next, the above comparison was repeated for different simulation settings (Fig. \ref{fig:various_settings}). To demonstrate that tGM-VAE can generalize to brain parcellations of different scales, the number of ROIs was varied between 10 and 50, which covered the typical range used in existing analyses of functional dynamics \cite{Allen12,Damarajua14,Taghia17,Nielsen15}. In all settings the two VAE-based approaches produced more accurate clustering, and tGM-VAE was the most accurate approach. This was also the case when the standard deviation of noise in synthetic rs-fMRI time series was varied between 0.05 to 1. Another important parameter (not relevant to clustering approaches) in the analysis of functional dynamics is the length of the sliding window for computing correlation matrices. Previous works often use a window size longer than the mean dwell time of connectivity states in order to reliably compute correlation values, but this strategy could potentially fail to differentiate dynamic connectivity patterns across neighboring states because the long window often covers multiple state transitions. While the analysis of window length is not the focus of the presented work, our experimental results (Fig. \ref{fig:various_settings}c) indicate that choosing a window size longer than the mean dwell time does not guarantee accurate clustering.

Note that the shallow neural networks tested here are a simplification choice and not a limitation of the method. Further exploration in the network structure would lead to better results for tGM-VAE. For instance, setting the dimension of latent space larger than 3 would produce higher accuracy for large correlation matrices (Fig. \ref{fig:various_settings}d).

\subsection{The NCANDA Dataset}

\begin{figure}[!t]
	\centering
    \includegraphics[width=1\linewidth]{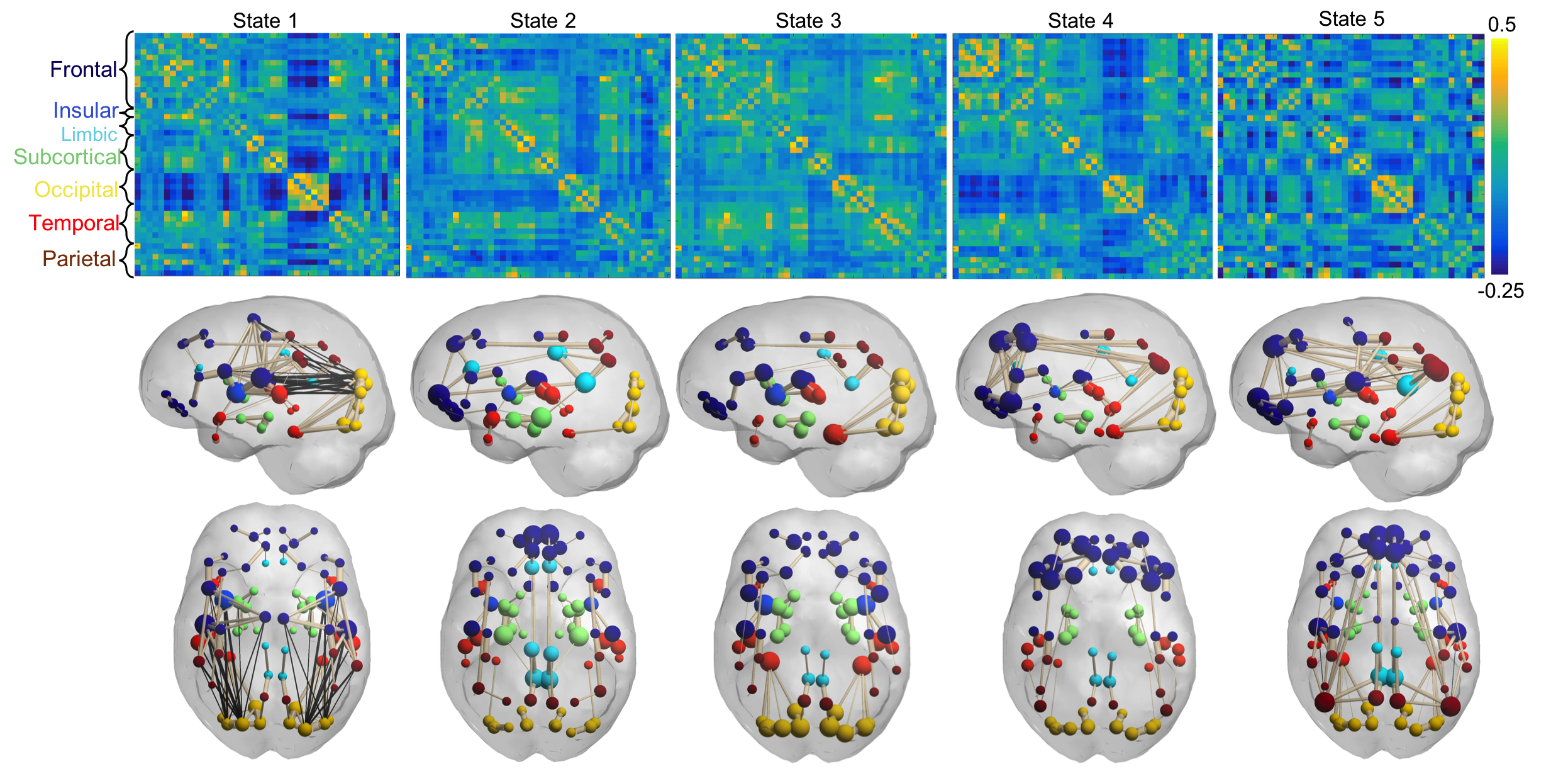}
	\caption{Functional connectivity patterns of 5 major states derived from the NCANDA rs-fMRI data. Top: mean correlation matrices; Bottom: Graph visualization of the mean correlation matrices. Node color corresponds to lobe names. Node size corresponds to sum of positive correlations associated with that node. White edges correspond to correlations $\geq0.25$ and black edges $\leq-0.25$. Edge thickness corresponds to absolute value of correlation.}
	\label{fig:ncanda}
\end{figure}

\begin{figure}[!t]
	\centering
    \includegraphics[width=1\linewidth]{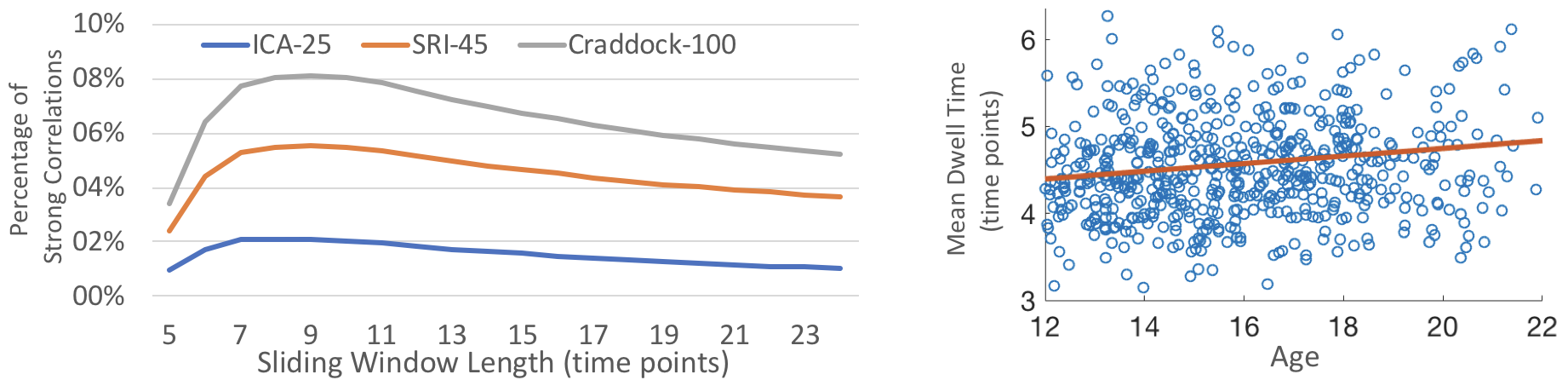}
	\caption{Left: The number of strong correlations (absolute value$\geq$0.5) depends on sliding window length but not on the number of ROIs in a parcellation. Right: For the NCANDA cohort, aging effect in the mean dwell time corrected for sex and scanner.}
	\label{fig:aging}
\end{figure}

We applied tGM-VAE to the rs-fMRI data of 593 normal adolescents (age 12-21; 284 boys and 309 girls) from the NCANDA study \cite{MO17} to investigate dynamic connectivity states in young brains. The rs-fMRI time series was preprocessed using the publicly available pipeline as described in the NCANDA study \cite{MO17}. For each subject, functional time series were extracted from 45 cerebral regions (averaged bilaterally) as defined by the sri24 atlas \cite{Rohlfing10}. Dynamic correlation matrices of size $45\times45$ were then derived for each subject based on a sliding-window approach \cite{Allen12} and improved by a linear shrinkage operation \cite{Chen10}. As mentioned, there is no consensus on the optimal length of the sliding-window. In the present work, we selected the length that produced the largest number of strong correlations (absolute value $\geq$ 0.5) to maximize the information contained in the training data. Our experiments suggest that the optimum was achieved at 10 time points (22s) regardless of the parcellation used to produce correlation matrices (Fig. \ref{fig:aging}). Afterwards, a total of 153587 matrices were derived for the entire cohort and clustered by tGM-VAE into 5 major states \cite{Damarajua14}. The dimension of the latent space was set to 6. Other parameters were set as in the synthetic experiments. Fig. \ref{fig:ncanda} shows the mean correlation matrices associated with the 5 major states detected by tGM-VAE and visualizes their graph structures. These 5 states correspond to well-known functional networks: auditory network (State 1), limbic and thalamo-striatal network (State 2), visual network (State 3), salience network (State 4) and the default mode network (State 5). 

Based on the clustering results, the state sequence was recovered for each subject and the mean dwell time over all states was computed. A group analysis was then performed to investigate the aging effect on the mean dwell time. First, sex and scanner-type were removed as confounding factors from mean dwell time using regression analysis \cite{MO17,Pfefferbaum17}. The residuals were then correlated with age, resulting in a significant positive correlation (one-tailed $p$=.0006, Fig \ref{fig:aging}). This age-related increase of mean dwell time could also be observed when the analysis was repeated with the dimension of latent space varying between 3 to 7. These results essentially indicate each connectivity state tends to persist longer in older adolescents, which converges with current concept of neurodevelopment that variation of dynamic functional connectivity declines with age \cite{Chen17}.

\section{Conclusion}
In this paper, we have presented a novel joint clustering and outlier-detection approach. Our model, tGM-VAE, introduces for the first time  a truncated Gaussian-mixture model in the variational autoencoder framework. This approach allows us to cluster data corrupted by noise, outliers and minor clusters of no interest to analysis. We used tGM-VAE to extract major functional connectivity states from resting-state fMRI scans and  characterize their dynamics. We showed that modeling latent representations of correlation matrices improves clustering accuracy compared to traditional Gaussian-mixture approaches and that our truncation strategy is useful in disentangling minor and major connectivity states. In the future, we will expand our framework to improve the modeling of state transitions.

\textbf{Acknowledgement} This research was supported in part by NIH grants
U24AA021697-06, AA005965, AA013521, AA017168.
%
%
%
\bibliographystyle{splncs}
\bibliography{ipmi}

\end{document}